\documentclass[twocolumn, a4paper]{article}
\usepackage{amssymb} 
\usepackage{graphicx} 
\usepackage[style=ieee]{biblatex} 

\usepackage{enumerate}
\usepackage{tabularray}
\usepackage[figuresright]{rotating} 
\usepackage{longtable}
\usepackage{caption}
\usepackage{subfig}
\usepackage{authblk}

\usepackage{array}
\usepackage{multirow}
\usepackage{ragged2e}
\usepackage{booktabs}

\graphicspath{{images/}} 
\usepackage[bookmarks=true, colorlinks, citecolor=black, linkcolor=black, urlcolor=black]{hyperref}
\title{\textbf{FaKnow: A Unified Library for Fake News Detection}}

\author[1]{Yiyuan Zhu}
\author[2]{Yongjun Li\thanks{Corresponding author: lyj@nwpu.edu.cn}}
\author[2]{Jialiang Wang}
\author[1]{Ming Gao}
\author[2]{Jiali Wei}

\affil[1]{School of Software, Northwestern Polytechnical University, Xi'an, Shaanxi 710072, China}
\affil[2]{School of Computer Science, Northwestern Polytechnical University, Xi'an,  Shaanxi 710072, China}

\date{}

\begin{document}
\maketitle
\begin{abstract}

Over the past years, a large number of fake news detection algorithms based on deep learning have emerged. However, they are often developed under different frameworks, each mandating distinct utilization methodologies, consequently hindering reproducibility. Additionally, a substantial amount of redundancy characterizes the code development of such fake news detection models. To address these concerns, we propose FaKnow, a unified and comprehensive fake news detection algorithm library. It encompasses a variety of widely used fake news detection models, categorized as content-based and social context-based approaches. This library covers the full spectrum of the model training and evaluation process, effectively organizing the data, models, and training procedures within a unified framework. Furthermore, it furnishes a series of auxiliary functionalities and tools, including visualization, and logging. Our work contributes to the standardization and unification of fake news detection research, concurrently facilitating the endeavors of researchers in this field. The open-source code and documentation can be accessed at https://github.com/NPURG/FaKnow and https://faknow.readthedocs.io, respectively.
\end{abstract}

\paragraph{Keywords}fake news detection toolkit, algorithms library, fake news detection framework

\section{INTRODUCTION}\label{sec:intro}

Nowadays, there is a notable abundance of fake news disseminated across popular social media platforms, such as Weibo\footnote{https://weibo.com/} and Twitter\footnote{https://twitter.com/}. This dissemination of unverified rumors, upon propagation, instigates substantial detrimental effects. Consequently, an imperative requirement emerges for tools and algorithms capable of discerning and classifying fake news, as the manual verification of such fake news proves prohibitively expensive and often time-consuming. Leveraging sophisticated neural network models, the effective application of deep learning within the realm of fake news identification becomes feasible. This, in turn, engenders improved identification of potential instances of fake news within the extensive corpus of social media data, thereby impeding the further diffusion of such rumors.

Different fake news detection algorithms that effectively leverage diverse characteristics associated with fake news are emerging alongside the rapid development of this area. However, it is important to note that each type of detection algorithm concentrates on distinct aspects, encompassing varying methodologies and applicable scenarios. While the processes of training, validating, and evaluating neural network models for such detection algorithms exhibit considerable similarities and closely interrelated components, they necessitate significant repetitive efforts demanding both time and labor. Besides, the implementation of these algorithms poses a significant challenge to academic researchers. Despite the availability of open-source code for most of these algorithms, the code is typically developed using different frameworks or platforms, each of which adopts a distinct dependency package for deep learning. The absence of a standardized specification for model construction and uniform calling interfaces hinders the versatile utilization of these codes. Furthermore, many of these open-source codes are confined to serving as system prototypes for algorithms, further complicating the independent adoption of these algorithms by researchers.

To address the aforementioned issues and facilitate researchers in creating and replicating fake news detection techniques, we propose a unified and comprehensive library based on PyTorch\footnote{https://pytorch.org/} for fake news detection algorithms called \textbf{FaKnow}(Fake Know). FaKnow integrates extensively recognized and widely used fake news detection algorithms from prominent academic journals and conferences in recent years. Furthermore, it incorporates a suite of essential workflows for model training and evaluation, encompassing a unified dataset format, optimal functionalities for training and evaluating models, intuitive visualization, logging capabilities, and efficient storage of model parameters. The following will provide an overview of this library's characteristics and capabilities from five perspectives.

\paragraph{Unified Framework}FaKnow provides a unified and standardized framework for various algorithms development encompassing data processing, model development, training and evaluation, and final result storage. The framework consists of three major modules, namely the data module, model module, and trainer module. FaKnow also incorporates a comprehensive set of components and functionalities that are commonly employed in fake news detection algorithms, effectively minimizing repetitive tasks and streamlining the algorithm development process.

\paragraph{Generic Data Structure}FaKnow incorporates various data formats to cater to different requirements arising from diverse tasks and scenarios. We have devised a standardized data format specifically tailored for content-based models, encompassing both text-based and multi-modal-based models. JSON(a lightweight data-interchange format with key-value pairs) is utilized as the file format for data input within the framework and FaKnow affords users the flexibility to customize the handling of different fields. By utilizing \textit{Dict} in Python for input batch data interaction between scripts, users can effortlessly retrieve values by referencing the predefined feature names embedded within the framework.

\paragraph{Diverse Models}FaKnow includes a collection of prominent fake news detection algorithms, encompassing a diverse array of content-based and social context-based models. These algorithms have been widely disseminated through esteemed academic conferences or journals, furnishing researchers with a comprehensive selection of options for both reproducing previous findings and employing them as baselines in their algorithmic research endeavors. The built-in models encompass a wide spectrum of perspectives, encompassing multi-modality, news propagation, and domain adaption. Moreover, our library transcends conventional algorithmic approaches, emphasizing contemporary and sought-after algorithms that have emerged in recent years.

\paragraph{Convenient Usability}FaKnow is constructed upon the foundation of PyTorch and incorporates encapsulation techniques to enhance its functionality. It alleviates the arduous tasks associated with common model training and evaluation processes. Additionally, it provides a range of auxiliary tools including result visualization, logging, and parameter saving, among others. These features effectively reduce code redundancy and facilitate seamless integration into workflows. Furthermore, the framework promotes configuratability by supporting the extraction of hyper-parameters from both configuration files and function arguments during model training. Despite the inclusion of its wrapper classes and functions, the framework maintains a gentle learning curve, enabling researchers familiar with PyTorch to swiftly commence their work.

\paragraph{Great Scalability}FaKnow contains classes for managing datasets, models, training, and evaluation. These classes are thoughtfully designed to abstract away intricate internal code logic and provide clear external call interfaces, thereby enhancing the overall extensibility of the framework. In addition to utilizing the built-in models bundled with the library, users seeking to fine-tune these models or develop new models can easily leverage the API(Application Programming Interface) exposed by the library. By inheriting existing classes and adhering to the specified guidelines, users can make use of the bulk of the framework's functionality while only needing to modify a minimal amount of code to meet their specific requirements.

In this paper, the research background and motivation are presented in Sec \ref{sec:intro}, with a focus on the proposed FaKnow library and its features. Sec \ref{sec:related} offers an in-depth review of the relevant research work about FaKnow. Then we outline the framework structure of this library and highlight its three significant modules in Sec \ref{sec:framework}. The reproducibility experiments conducted on integrated models in FaKnow are represented in Sec \ref{sec:experiment}. Sec \ref{sec:usage} encompasses a concise presentation of the library's basic usage, complemented by illustrative code examples. Lastly, Sec \ref{sec:conclusion} furnishes a comprehensive summary of our work, concluding with a forward-looking analysis of its prospects.

\section{RELATED WORK}\label{sec:related}
In this section, we briefly review the work related to the proposed fake news detection algorithms library. We mainly focus on the following two topics: fake news detection and deep learning algorithms libraries.

\subsection{Fake News Detection}\label{sec:fnd}
The current landscape of fake news detection algorithms encompasses two major categories: content-based, and social context-based.

Content-based detection algorithms focus mainly on the content in the post, and can effectively conduct early detection of fake news. The more typical detection algorithms\cite{chengVRoCVariationalAutoencoderaided2020, vaibhavSentenceInteractionsMatter2019, yuConvolutionalApproachMisinformation2017a} take the text of the post as the model input, obtain the word embedding by Word2Vec\cite{w2v} or BERT\cite{bert}, and extract the text features for detection by TextCNN\cite{textcnn}, LSTM\cite{lstm}, etc. Meanwhile, the images in the posts can also provide crucial information, thus the multi-modal approach is widely employed. SpotFake\cite{spotfake} retrieves image features via pre-trained VGG\cite{vgg} and feeds them into the classifier after simply concatenating them with text features. MCAN\cite{mcan} fuses spatial domain features, frequency domain features, and text features via multiple stacks of Co-Attention layers. HMCAN\cite{hmcan} extracts text features using a hierarchical encoding network and fuses them with image features via a complex multi-modal contextual Transformer. There is also some research like SAFE\cite{safe} and EM-FEND\cite{emfend} focusing on the consistency between different modalities, and information such as the cosine similarity between texts and images is used to assist in discriminating fake news. In addition, posts in different news domains often have their characteristics in terms of text and other aspects, so some studies concentrate on the news domains and extract domain features to improve the generalization ability of the model in detecting posts from various domains. EANN\cite{eann} proposes a novel event adversarial neural network framework that can learn transferable features for unseen events via the event discriminator removing the event-specific features. MDFEND\cite{mdfend} utilizes the domain gate to aggregate multiple representations extracted by a mixture of experts for multi-domain fake news detection.

Social context-based detection algorithms usually treat the post in a social network rather than an isolated individual and utilize various information from the social network such as user profiles, comments on the post, news propagation, etc. for detection. GCAN\cite{gcan} builds a user propagation graph with re-tweeting users as nodes and user profiles as nodes representation. GCN\cite{gcn} is used to extract the features of the graph, which are eventually fused with the features of the tweets to detect fake news. BiGCN\cite{bigcn} constructs top-down and bottom-up post-propagation graphs, respectively, and extracts their respective features with GCN\cite{gcn} and eventually concatenates the two to feed the classifier. UPFD\cite{upfd} takes into account the user preferences and uses the tweets history by the publisher of the post to be detected as an endogenous factor, and constructs post-propagation graphs as an exogenous factor. These two factors are combined to determine the credibility of the post. Fang\cite{fang} constructs a heterogeneous graph with three types of nodes: user, news articles, and news source, extracts features through GraphSage\cite{graphsage} and Bi-LSTM\cite{lstm} and introduces three loss functions to optimize the model, namely Proximity Loss, Stance Loss, and Fake News Loss.

\subsection{Specific Algorithms Libraries}
Currently, there are several specially designed open-source deep learning algorithm libraries in a variety of application fields. The majority of these algorithm libraries integrate general models in their specific research areas, cover many processes like model training, and offer a convenient calling interface, allowing users to get started quickly. RecBole\cite{recbole} is an open-source recommendation system library, which provides a unified framework to develop and reproduce recommendation algorithms and is also useful for standardizing the evaluation protocol of recommendation algorithms. It integrates 73 recommendation models on 28 benchmark datasets, covering the categories of general recommendation, sequential recommendation, context-aware recommendation, and knowledge-based recommendation. MMRec\cite{mmrec} is a library of recommender algorithms dedicated to multi-modal recommendations, which simplifies and canonicalizes the process of implementing and comparing multi-modal recommendation models and provides a unified and configurable arena that can minimize the effort in implementing and testing multi-modal recommendation models. It enables multi-modal models, ranging from traditional matrix factorization to modern graph-based algorithms, capable of fusing information from multiple modalities simultaneously. LibCity\cite{libcity} includes all the necessary steps or components related to traffic prediction into a systematic pipeline with various datasets, mechanisms, models, and utilities and covers four mainstream tasks, including traffic speed prediction, traffic flow prediction, on-demand service prediction, and trajectory next location prediction. Transformers\cite{transformers} is a library consisting of carefully engineered state-of-the-art transformer-based architectures under a unified API. It supports the distribution and usage of a wide variety of pre-trained models which facilitate users to perform tasks on different modalities such as text, vision, and audio.

Although many practical and user-friendly open-source algorithm libraries have appeared in some research areas like recommender systems, and many researchers continue to focus on the development and research of algorithm libraries, there has been a dearth of such a unified open-source algorithm library in fake news detection, which has brought a great deal of inconvenience for the research in this area.

\section{LIBRARY FRAMEWORK}\label{sec:framework}

\begin{figure}[!htbp]
    \centering
    \includegraphics[width=0.49\textwidth]{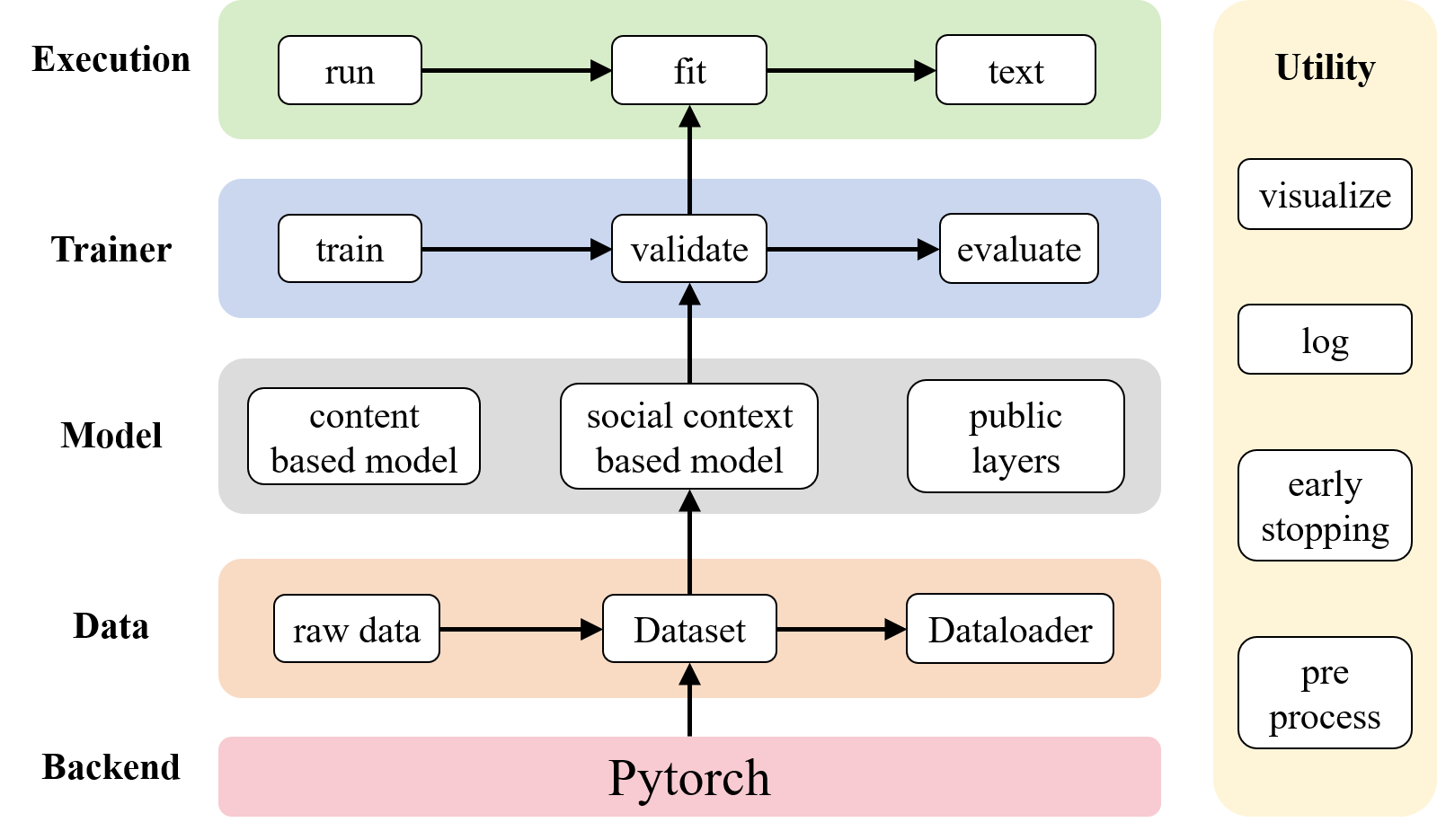}
    \caption{FaKnow framework}
    \label{fig:framework}
\end{figure}

The overall framework of the FaKnow library is shown in Figure \ref{fig:framework}. FaKnow takes PyTorch as its backend and consists of five modules. The data module is in charge of the data input into the model, including unified dataset format and data processing, etc. The model module incorporates several popular fake news detection models as well as some frequently used neural network components. These models are trained, validated, and evaluated by the trainer module. The execution module is used to execute the algorithms in the library, which organizes the other modules logically and provides a convenient way for users to run the built-in algorithms quickly, allowing users to pass in the required arguments from the configuration file or keywords dictionary. The utility module contains a number of useful utilities needed to run the program, including data visualization, logging, and early stopping. The data, model, and trainer modules are the core modules of the library and are introduced in detail in the following sections.

\subsection{Data Module}
Along with certain frequently used data processing functions, the data module includes all the \textit{pytorch.Dataset} classes required for the models integrated into FaKnow. Users can additionally expand or create new dataset classes to handle various scenarios.

\begin{figure*}[!htb]
    \centering
    \includegraphics[width=1\textwidth]{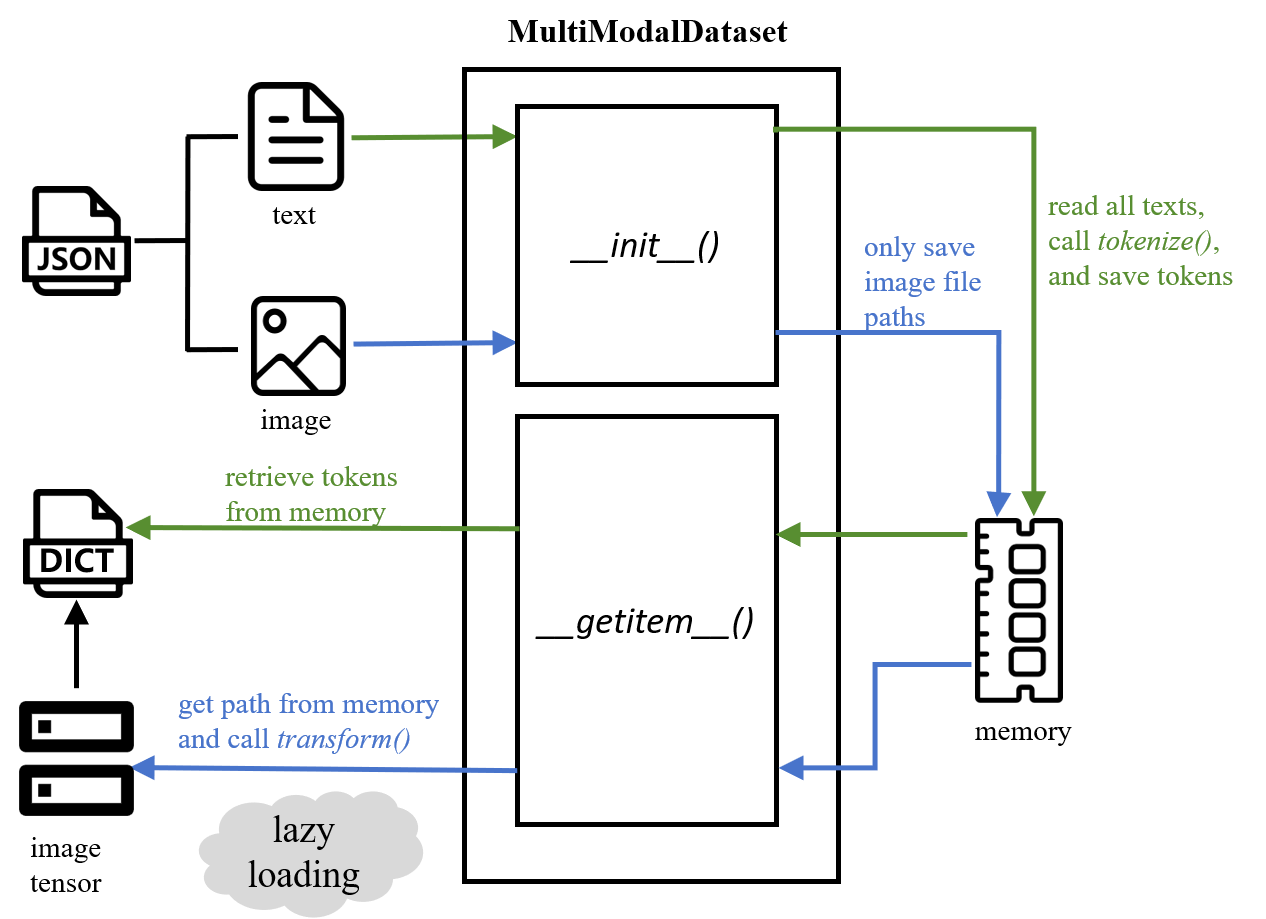}
    \caption{Multi-modal data processing}
    \label{fig:data}
\end{figure*}

\subsubsection{Dict batch data}\label{sec:dict_batch_data}
To make the model training uniform, FaKnow takes \textit{Dict} in Python, a data structure in the form of key-value pairs, as the format of the batch input data and uses the feature name as the key and the corresponding \textit{pytorch.Tensor} as the value, which allows the user to easily refer to their names to obtain the corresponding features. These dataset classes included in the data module all inherit from \textit{pytorch.Dataset}, and can be iteratively traversed by the \textit{\_\_getitem\_\_} method\footnote{To distinguish, we refer to functions in a class as methods according to the terminology of object-oriented programming.} to obtain the above \textit{Dict} data.

\subsubsection{Data structure for content-based models}\label{sec:dataset}

For content-based detection algorithms, we also design specialized data structures for storing these features from content. Since most fake news datasets are posts crawled from social platforms, and to fit the way of referencing data through feature names mentioned above, FaKnow adopts JSON as the format of the raw data file. All sample entities are recorded as an array in the JSON file, and each sample is a JSON object comprising key-value pairs.

For textual and multi-modal datasets, respectively, FaKnow involves built-in \textit{TextDataset} and \textit{MultiModalDataset} classes, with the \textit{MultiModalDataset} class inheriting from \textit{TextDataset}. These two dataset classes can extract samples from the aforementioned JSON data file, which includes fields like texts, image file paths, labels, etc., and offer users the flexibility to customize the processing of different fields according to their specific requirements.

Take \textit{MultiModalDataset} as an example, as shown in Figure \ref{fig:data}, it can handle multi-modal data that includes both texts and images. Users simply need to pass in the text processing function \textit{tokenize} and the image processing function \textit{transform}, as well as the names of the text and image fields in the JSON file. Thus, the inherent \textit{\_\_getitem\_\_} method of \textit{MultiModalDataset} facilitates the retrieval of a \textit{Dict} object that comprises both text and image data. To ensure flexibility, both the image processing logic code (containing operations like reading pixels) in \textit{transform} and the text processing logic code (including word segmentation and other operations) in \textit{tokenize} can be customized by users.

\begin{figure}[!htbp]
    \centering
    \includegraphics[width=0.49\textwidth]{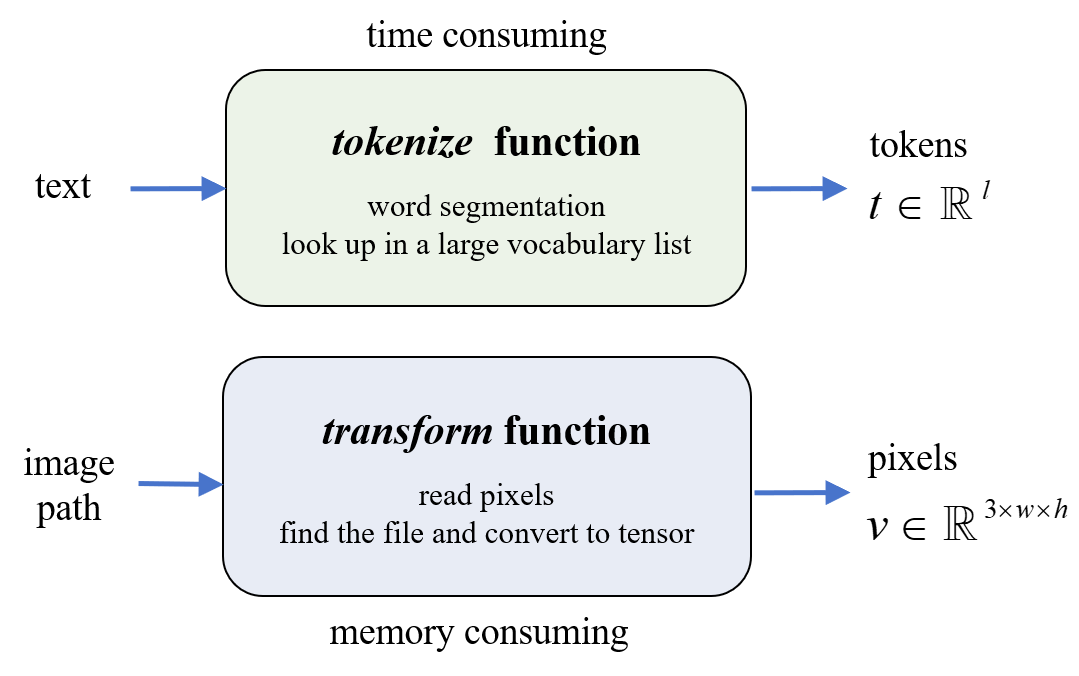}
    \caption{Tokenize and transform function}
    \label{fig:tokenize_transform}
\end{figure}

Figure \ref{fig:tokenize_transform} shows the difference between text processing and image processing. For text, the process of word segmentation necessitates substantial time due to the requirement of referencing an extensive vocabulary list. Consequently, the resulting token id sequence, often represented as a one-dimensional array denoted as $t \in \mathbb{R} ^l$(where $l$ is the length of the token sequence), occupies a relatively insignificant amount of memory. In light of these considerations, \textit{MultiModalDataset} effectively addresses the text-related operations by invoking the user-provided \textit{tokenize} function during initialization. 
When accessing the data within \textit{MultiModalDataset}, the corresponding token id sequence associated with a specific sample index can be directly obtained. 

On the contrary, the process of acquiring pixel values from an image is often less time-consuming. However, the storage of image data with RGB channels typically necessitates the allocation of a three-dimensional array denoted as $v \in \mathbb{R}^{3\times w\times h}$(where $w$ and $h$ are the width and height of the image respectively), which consumes a significant amount of memory. Thus, we employ a strategy that prioritizes a \textbf{time-space tradeoff} when it comes to image processing. For images, the \textit{MultiModalDataset} is designed to initialize by storing image file paths. During the traversal, the \textit{transform} function provided by users is invoked to fetch the image data into memory. This implementation employs a lazy loading approach, wherein image data is read into memory only when necessary, thus conserving significant amounts of memory space. 


\subsection{Model Module}
Based on the aforementioned data module, we have systematically categorized all fake news detection models within FaKnow into the model module, while simultaneously providing a cohesive and standardized calling interface.

\subsubsection{Integrated models}
Table \ref{table:models} illustrates models sourced from recent publications in esteemed conferences and journals, including AAAI, SIGIR, IJCAI, ACL, KDD, and others, into the comprehensive FaKnow library. The meticulous selection of integrated models aimed to maximize heterogeneity and furnish users with a wide array of choices to tackle an assortment of tasks. More specifically, the chosen papers encompass two major categories: content-based and social context-based, thereby encompassing state-of-the-art methodologies spanning diverse technologies, such as multi-modality, domain adaptation, and graph neural networks. Besides, our library encompasses universal classification algorithms, including TextCNN\cite{textcnn} for text classification, as well as GCN\cite{gcn}, GAT\cite{gat}, and GraphSAGE\cite{graphsage} for graph classification. Additionally, in a concerted effort to alleviate code redundancy, certain neural network components frequently employed in fake news detection algorithms are also furnished as standalone entities within the data module. Examples of such components include the TextCNN\cite{textcnn} layer, facilitating the extraction of text features, the Discrete Cosine Transform layer, integral for extracting image frequency-domain features, and the Co-Attention layer, instrumental in multi-modal features fusion. These components, apart from being utilized by built-in models, can be readily reused by users for the development of new models.

\begin{table}
\caption{Integrated models}
\label{table:models}
\centering
\begin{tblr}{
  row{even} = {c},
  row{3} = {c},
  row{5} = {c},
  row{7} = {c},
  row{9} = {c},
  row{11} = {c},
  row{13} = {c},
  row{15} = {c},
  row{17} = {c},
  row{19} = {c},
  row{21} = {c},
  row{23} = {c},
  cell{1}{1} = {c},
  cell{1}{2} = {c},
  cell{1}{3} = {c},
  cell{2}{1} = {r=11}{},
  cell{13}{1} = {r=11}{},
  hline{1,24} = {-}{0.08em},
  hline{2,13} = {-}{},
}
\textbf{category}                                   & \textbf{model} & \textbf{venue} & \textbf{year} \\
{\textbf{content}\\\textbf{based}}                  & TextCNN\cite{textcnn}        & EMNLP          & 2014          \\
                                                    & EANN\cite{eann}           & KDD            & 2018          \\
                                                    & SpotFake\cite{spotfake}       & BigMM          & 2019          \\
                                                    & SAFE\cite{safe}           & PAKDD          & 2020          \\
                                                    & MDFEND\cite{mdfend}         & CIKM           & 2021          \\
                                                    & MCAN\cite{mcan}            & ACL            & 2021          \\
                                                    & HMCAN\cite{hmcan}           & SIGIR          & 2021          \\
                                                    & MFAN\cite{mfan}            & IJCAI          & 2022          \\
                                                    & ENDFN\cite{endfn}           & SIGIR          & 2022          \\
                                                    & M3FEND\cite{m3fend}          & TKDE           & 2022          \\
                                                    & CAFE\cite{cafe}            & WWW            & 2022          \\
{\textbf{social}\\\textbf{context}\\\textbf{based}} & GCN\cite{gcn}             & ICLR           & 2017          \\
                                                    & GraphSAGE\cite{graphsage}      & NeurIPS        & 2017          \\
                                                    & GAT\cite{gat}             & ICLR           & 2018          \\
                                                    & GCNFN\cite{gcnfn}           & arXiv          & 2019          \\
                                                    & BIGCN\cite{bigcn}           & AAAI           & 2020          \\
                                                    & FANG\cite{fang}            & CIKM           & 2020          \\
                                                    & UPFD\cite{upfd}            & SIGIR          & 2021          \\
                                                    & GNNCL\cite{gnncl}           & ICANN          & 2021          \\
                                                    & DUDEF\cite{dudef}           & WWW            & 2021          \\
                                                    & EBGCN\cite{ebgcn}           & ACL            & 2021          \\
                                                    & TrustRD\cite{trustrd}         & CIKM           & 2023          
\end{tblr}
\end{table}

\subsubsection{Unified interface} \label{sec:interface}

\begin{figure}[!htbp]
    \centering
    \includegraphics[width=0.38\textwidth]{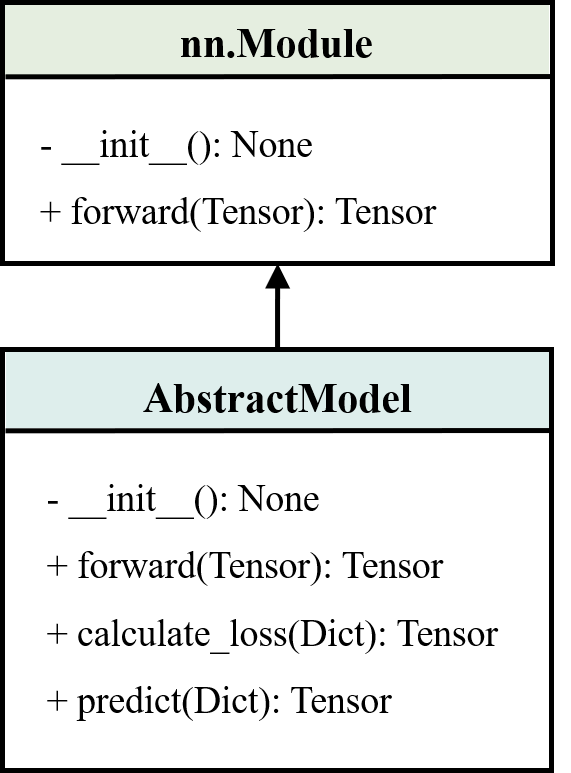}
    \caption{UML class diagram}
    \label{fig:abstractmodel}
\end{figure}

We designed an abstract class called \textit{AbstractModel} which inherits from the \textit{nn.Module} class in PyTorch and serves as the parent class for all models incorporated in FaKnow. Therefore, the underlying implementation logic remains congruent with that of PyTorch, necessitating the overriding of the \textit{\_\_init\_\_} and \textit{forward} methods for model initialization and forward propagation, respectively. Nevertheless, aiming to establish a standardized interface for invoking the models, there are two new methods in \textit{AbstractModel}, namely \textit{calculate\_loss} and \textit{predict}, which should be implemented by all models in the library. Figure \ref{fig:abstractmodel} illustrates the UML class diagram of \textit{AbstractModel}.

\begin{figure}[!htbp]
    \centering
    \includegraphics[width=0.49\textwidth]{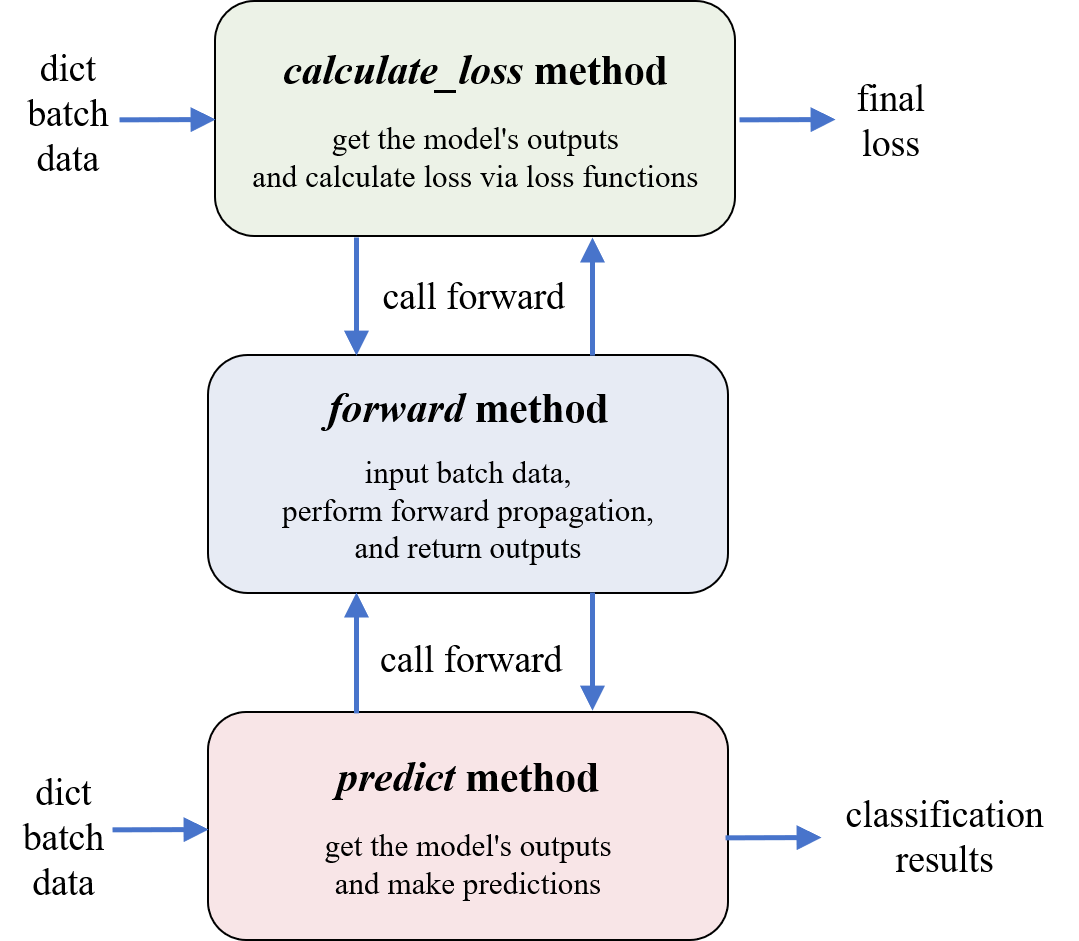}
    \caption{\textit{calculate\_loss} and \textit{predict} methods}
    \label{fig:interface}
\end{figure}

As shown in Figure \ref{fig:interface}, the newly introduced \textit{calculate\_loss} and \textit{predict} both accept the \textit{Dict} data presented in the data module as input for batched samples. Subsequently, these two methods invoke the \textit{forward} method to obtain outputs of the final layer in the model. The former computes the loss through the designated loss function and subsequently returns the computed loss which plays a pivotal role in parameter updates through back-propagation during the training phase. On the other hand, the latter returns the classification outcome of the model's prediction of fake news. This aids in model inference and evaluation.

Notably, irrespective of whether the model exhibits multiple outputs at the final layer or possesses a final loss formed through cumulative losses, the model's invocation during the training and testing phases remains unified. For users who want to develop new models utilizing FaKnow, the task merely entails overriding these two interface methods, alleviating concerns regarding intricate details about model invocation. Further insights into the implementation details of new models are elaborated upon in Sec \ref{sec:develop}.

\subsection{Trainer Module}
To address the imperative of streamlining the training process for diverse models within FaKnow and alleviate the burden of repetitive work, we have ingeniously devised the trainer module. Positioned after the aforementioned data module and model module, this module assumes the crucial responsibility of feeding the data into the model in batches.

It encompasses a plethora of essential functionalities, including but not limited to model training, validation, testing, saving, logging of training progress, and visualization thereof. These functionalities are harmoniously encapsulated within the class called \textit{Trainer}, rendering it impervious to users' intricate internalities. By simply specifying the desired model, optimization algorithm, and other hyper-parameters, users can effortlessly expedite the training process. The trainer module further embraces advanced settings, such as gradient clipping, the learning rate scheduler, and early stopping, to accommodate divergent circumstances and demands.

\begin{figure}[!htbp]
    \centering
    \includegraphics[width=0.49\textwidth]{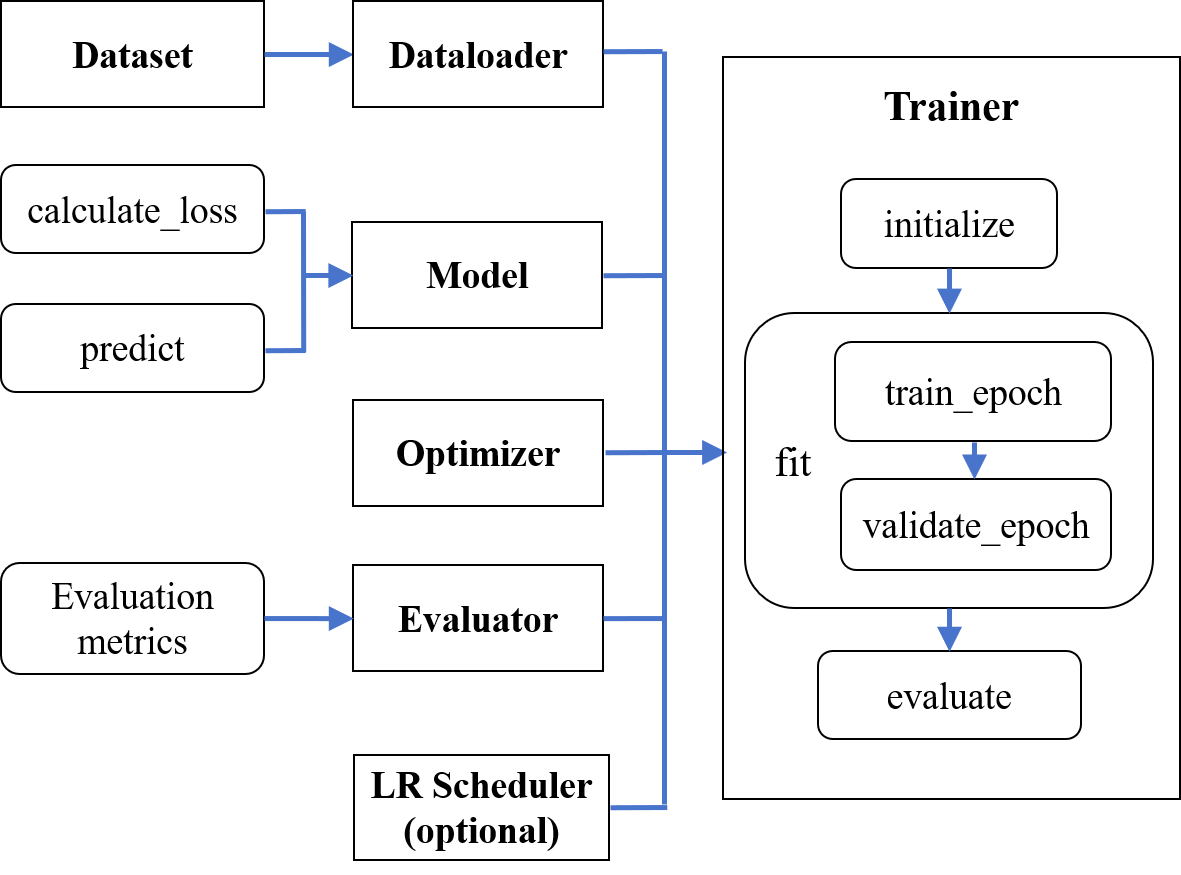}
    \caption{Workflow in Trainer}
    \label{fig:trainer}
\end{figure}

\subsubsection{Training process}

Figure \ref{fig:trainer} illustrates the workflow in \textit{Trainer}. During the training process of the \textit{Trainer}, adherence to established deep learning model training protocols is maintained. This involves partitioning the data into three distinct subsets: the training set, the validation set, and the test set. Following each training epoch, model validation is performed, while the model is tested only upon completion of all training epochs. Classification metrics including accuracy, precision, recall, and F1 score are subsequently computed.

Upon initialization of the \textit{Trainer}, the user is required to provide fundamental arguments like the intended model, the optimizer for parameter optimization, and the evaluation metrics within the \textit{\_\_init\_\_} method. Subsequently, the trainer allows for the invocation of the \textit{fit} method to commence model training. This method accepts arguments such as training and validation data, as well as the number of training epochs. In each epoch, it employs the \textit{train\_epoch} and \textit{validate\_epoch} methods in \textit{Trainer} to respectively train and validate the model. It also provides real-time updates on the training loss and the classification performance of the model on the validation set, thereby enabling continuous monitoring of the model's training progress. Additionally, to provide users with maximal flexibility, \textit{fit} exclusively focuses on training the model in the absence of validation data. After the completion of training, the \textit{evaluate} method can be invoked to assess the model's performance on the test set, which takes the test data as input and returns classification metrics as output.

\subsubsection{Auxiliary functionalities}
To enhance user convenience, the trainer module not only facilitates fundamental model training but also encompasses supplementary functionalities. Regarding model training, the \textit{Trainer} seamlessly integrates numerous advanced configurations. 

\begin{itemize}
    \item gradient clipping: is invoked during the training of every batch of data to avert gradient explosion
    \item learning rate scheduler: adjusts the learning rate as instructed by the user after each epoch of training and validation to mitigate over-fitting.
    \item early stopping: determines whether to prematurely terminate the training process based on the model's performance on the validation set in each iteration and save the model with the best performance on the validation set, alongside corresponding evaluation results from all iterations.
    \item logging: shows training loss and validation results in the console and saves them to a local log file.
    \item visualization: visualizes the fluctuation curves encompassing training loss and validation results (including accuracy and other metrics) from each iteration with TensorBoard\footnote{https://www.tensorflow.org/tensorboard}.
    \item to device: moves data and model to the specific device like Cuda or CPU.
\end{itemize}


If the aforementioned features fail to cater to user requirements adequately, users may opt to inherit the \textit{Trainer} class, thereby leveraging existing code as far as possible for developing new functionalities.

\section{EXPERIMENTS}\label{sec:experiment}


\begin{table*}
\centering
\caption{Reproducibility experiments result}
\label{table:reproduce}
\resizebox{.95\linewidth}{!}{
\begin{tblr}{
  cells = {c},
  cell{1}{1} = {r=2}{},
  cell{1}{2} = {r=2}{},
  cell{1}{3} = {r=2}{},
  cell{1}{4} = {c=5}{},
  cell{3}{1} = {r=2}{},
  cell{3}{2} = {r=2}{},
  cell{5}{1} = {r=2}{},
  cell{5}{2} = {r=2}{},
  cell{7}{1} = {r=2}{},
  cell{7}{2} = {r=2}{},
  cell{9}{1} = {r=2}{},
  cell{9}{2} = {r=2}{},
  cell{11}{1} = {r=2}{},
  cell{11}{2} = {r=2}{},
  cell{13}{1} = {r=2}{},
  cell{13}{2} = {r=2}{},
  cell{15}{1} = {r=2}{},
  cell{15}{2} = {r=2}{},
  cell{17}{1} = {r=2}{},
  cell{17}{2} = {r=2}{},
  cell{19}{1} = {r=2}{},
  cell{19}{2} = {r=2}{},
  cell{21}{1} = {r=2}{},
  cell{21}{2} = {r=2}{},
  cell{23}{1} = {r=2}{},
  cell{23}{2} = {r=2}{},
  cell{25}{1} = {r=2}{},
  cell{25}{2} = {r=2}{},
  cell{27}{1} = {r=2}{},
  cell{27}{2} = {r=2}{},
  cell{29}{1} = {r=2}{},
  cell{29}{2} = {r=2}{},
  cell{31}{1} = {r=2}{},
  cell{31}{2} = {r=2}{},
  cell{33}{1} = {r=2}{},
  cell{33}{2} = {r=2}{},
  cell{35}{1} = {r=2}{},
  cell{35}{2} = {r=2}{},
  cell{37}{1} = {r=2}{},
  cell{37}{2} = {r=2}{},
  hline{1,39} = {-}{0.08em},
  hline{2} = {4-8}{},
  hline{3} = {-}{},
}
\textbf{model} & \textbf{dataset}          & \textbf{results} & \textbf{metrics} &                    &                 &             &              \\
               &                           &                  & \textbf{acc}     & \textbf{precision} & \textbf{recall} & \textbf{f1} & \textbf{auc} \\
EANN\cite{eann}           & Weibo17\cite{weibo17}                   & original         & 0.827            & 0.847              & 0.812           & 0.829       & -            \\
               &                           & ours             & 0.800            & 0.800              & 0.790           & 0.790       & -            \\
SpotFake\cite{spotfake}       & TwitterMediaEval16\cite{mediaeval1516}        & original         & 0.777            & 0.791              & 0.753           & 0.760       & -            \\
               &                           & ours             & 0.769            & 0.765              & 0.866           & 0.812       & -            \\
SAFE\cite{safe}             & Politifact\cite{fakenewsnet}                & original         & 0.874            & 0.889              & 0.903           & 0.896       & -            \\
               &                           & ours             & 0.791            & 0.836              & 0.796           & 0.816       & -            \\
MDFEND\cite{mdfend}         & Weibo21\cite{mdfend}                   & original         & -                & -                  & -               & 0.913       & -            \\
               &                           & ours             & -                & -                  & -               & 0.912       & -            \\
MCAN\cite{mcan}            & Weibo17\cite{weibo17}                   & original         & 0.899            & 0.898              & 0.899           & 0.899       & -            \\
               &                           & ours             & 0.873            & 0.919              & 0.832           & 0.873       & -            \\
HMCAN\cite{hmcan}          & Weibo17\cite{weibo17}                   & original         & 0.885            & 0.888              & 0.885           & 0.885       & -            \\
               &                           & ours             & 0.816            & 0.800              & 0.841           & 0.820       & -            \\
MFAN\cite{mfan}           & CED\cite{ced}                       & original         & 0.889            & 0.889              & 0.881           & 0.883       & -            \\
               &                           & ours             & 0.888            & 0.896              & 0.870           & 0.879       & -            \\
ENDEF\cite{endfn}          & WeiboNEP\cite{nep}                  & original         & 0.806            & -                  & -               & 0.731       & 0.849        \\
               &                           & ours             & 0.846            & -                  & -               & 0.802       & 0.877        \\
M3FEND\cite{m3fend}         & FakeNewsNet\cite{fakenewsnet}\&MMCovid\cite{mmcovid}        & original         & 0.897            & -                  & -               & 0.851       & 0.934        \\
               &                           & ours             & 0.921            & -                  & -               & 0.920       & 0.976        \\
CAFE\cite{cafe}           & TwitterMediaEval15\cite{mediaeval1516}        & original         & 0.806            & 0.806              & 0.806           & 0.806       & -            \\
               &                           & ours             & 0.840            & 0.868              & 0.812           & 0.839       & -            \\
GCNFN\cite{gcnfn}          & Politifact\cite{fakenewsnet}                & original         & 0.832            & -                  & -               & 0.836       & -            \\
               &                           & ours             & 0.850            & -                  & -               & 0.889       & -            \\
BIGCN\cite{bigcn}          & Twitter16\cite{twitter1516}                 & original         & 0.880            & -                  & -               & 0.879       & -            \\
               &                           & ours             & 0.868            & -                  & -               & 0.854       & -            \\
FANG\cite{fang}           & FakeNewsNet\cite{fakenewsnet}\&PHEME\cite{pheme}\&TwitterMa\cite{twitterma} & original         & -                & -                  & -               & -           & 0.751        \\
               &                           & ours             & -                & -                  & -               & -           & 0.766        \\
UPFD\cite{upfd}           & Politifact\cite{fakenewsnet}                & original         & 0.846            & -                  & -               & 0.846       & -            \\
               &                           & ours             & 0.833            & -                  & -               & 0.817       & -            \\
GNNCL\cite{gnncl}          & Politifact\cite{fakenewsnet}                & original         & 0.629            & -                  & -               & 0.622       & -            \\
               &                           & ours             & 0.660            & -                  & -               & 0.714       & -            \\
DUDEF\cite{dudef}          & Weibo20\cite{dudef}                   & original         & 0.855            & -                  & -               & 0.855       & -            \\
               &                           & ours             & 0.865            & -                  & -               & 0.893       & -            \\
EBGCN\cite{ebgcn}          & Twitter16\cite{twitter1516}                 & original         & 0.915            & -                  & -               & 0.910       & -            \\
               &                           & ours             & 0.837            & -                  & -               & 0.820       & -            \\
TrustRD\cite{trustrd}        & Twitter15\cite{twitter1516}                 & original         & 0.931            & -                  & -               & 0.927       & -            \\
               &                           & ours             & 0.924            & -                  & -               & 0.846       & -            
\end{tblr}}
\end{table*}


To ensure the correctness of the integrated models in FaKnow and enable reproducibility, we conducted several experiments on different datasets to compare results with those reported in the original paper. However, for classification models like TextCNN\cite{textcnn} and GCN\cite{gcn}, which are applied to general tasks, we did not perform reproducibility experiments on the fake news dataset.

\subsection{Implementation Details}
In our experiments, datasets, evaluation metrics, and all hyper-parameters, including learning rate, number of training epochs, and batch size, were strictly aligned with those specified in the open-source code from the original paper. In instances where the original paper did not report specific metrics, we utilized blank characters in the corresponding table cells.

To align with the original paper, MDFEND\cite{mdfend}, BiGRU, and GraphSage\cite{graphsage} are employed as base models for ENDFN\cite{emfend}, DUDEF\cite{dudef}, and UPFD\cite{upfd} frameworks, respectively, following the paper's methodology for comparison. Given the unavailability of open-source code from the original authors of GCNFN\cite{gcnfn} and GNNCL\cite{gnncl}, we utilized datasets and source code of these two models released as baselines from the UPFD\cite{upfd} paper, replicating their methodology for reproducibility experiments. The input features for the UPFD-GraphSage\cite{upfd}, GNNCL\cite{gnncl}, and GCNFN\cite{gcnfn} models are bert, profile, and content, respectively. Furthermore, due to the lack of data pre-processing code or processed data files from the authors of HMCAN\cite{hmcan}, we had to develop our own code to process the raw text and images in the dataset, following the implementation guidelines outlined in the paper. In addition, some broken image files in the dataset uploaded by the authors of SAFE\cite{safe} were removed, and the remaining intact dataset was used to train the model.

\subsection{Results Analysis}
Table \ref{table:reproduce} displays the results of our reproducibility experiments on built-in models in FaKnow, comparing them with evaluation metrics from the original paper. Notably, in the majority of cases, our results align closely with the original paper's metrics, indicating a minimal difference of only 1 to 2 percent, well within the acceptable margin of error.

Regarding content-based detection algorithms, exemplified by MFAN\cite{mfan}, the smallest disparity between our reproduced results on CED\cite{ced} and those in the original paper is evident in the accuracy metric, exhibiting a mere 0.001 gap. Meanwhile, the most significant variation is found in the recall metric, with a difference of only 0.11. For social context-based models like UPFD\cite{upfd}, the reproduced accuracy and f1 scores on Polifact\cite{fakenewsnet} are 0.833 and 0.817, respectively, again pretty close to the experimental outcomes in the original paper. Moreover, the results of the three models ENDEF\cite{endfn}, CAFE\cite{cafe}, and GNNCL\cite{gnncl} even surpass the original article's outcomes by approximately 3 percentage points.

However, a slight variation between the reproduced results and those in the original paper was observed in certain models, potentially attributed to nuanced differences in experimental conditions. These disparities present an opportunity for further study, offering insights to enhance our understanding of the model and potentially improve the current methodology.

Regarding HMCAN\cite{hmcan}, our reproduced accuracy, precision, recall, and f1 scores stand at 0.816, 0.8, 0.841, and 0.82 respectively, exhibiting a discrepancy of 4 to 8 percentage points when compared to the results outlined in the original paper. This variation may stem from disparities in our data pre-processing procedures in contrast to the approach employed by the authors. Meanwhile, the reproduction results on TwitterMediaEval15\cite{mediaeval1516} also demonstrate a reduction by several percentage points in comparison to the original results but closely align with the reproduced outcomes of this model on these two datasets in this paper\cite{csfnd}. 

In our experiments, the SAFE\cite{safe} model exhibited diminished performance during training with a crippled dataset compared to the metrics outlined in the original paper. Furthermore, generating text descriptions corresponding to images significantly impacts the model's effectiveness. Conversely, the pre-trained model ShowAndTell\cite{showandtell} to abstract the content of images is notably influenced by the dataset used for its training, often introducing biases when transposed to new tasks or datasets.

Additionally, the results of EBGCN\cite{ebgcn} exhibit an approximately 8-percentage-point decrease, potentially arising from a deficiency in the unsupervised Edge-wise Consistency module which is designed for unlabelled potential edge prediction. This module may inadequately learn the latent relationships between nodes, consequently impacting the model's training.

Regardless, apart from the mentioned exceptions, the reproduction experiment results are very close to the original paper across various evaluation metrics such as accuracy, demonstrating the effectiveness of the reproduced models in our library.

\section{USAGE EXAMPLES}\label{sec:usage}

In this section, we will briefly introduce how to use FaKnow and give some examples, which are unfolded in two main parts, running the models built into the library and developing new models based on it. For more usage details, please refer to our documentation.

\subsection{Run Integrated Models}
\subsubsection{Quick start}
FaKnow offers users a convenient way to expedite their engagement with the system by furnishing two key functions: \textit{run} and \textit{run\_from\_yaml}. These functions serve as comprehensive encapsulations of all the requisite processes entailed in model training and evaluation. Users are solely tasked with specifying the model name alongside its associated arguments, facilitating a quick program initiation. The former function accepts keyword arguments, encompassing input facets like datasets, model initialization arguments, and training hyper-parameters. Conversely, the latter leverages a YAML(a human-readable data serialization format that is often used for configuration files with a markup language) file to extract the essential arguments required for program execution.

\begin{figure*}[!htbp]
    \centering
    \includegraphics[width=1\textwidth]{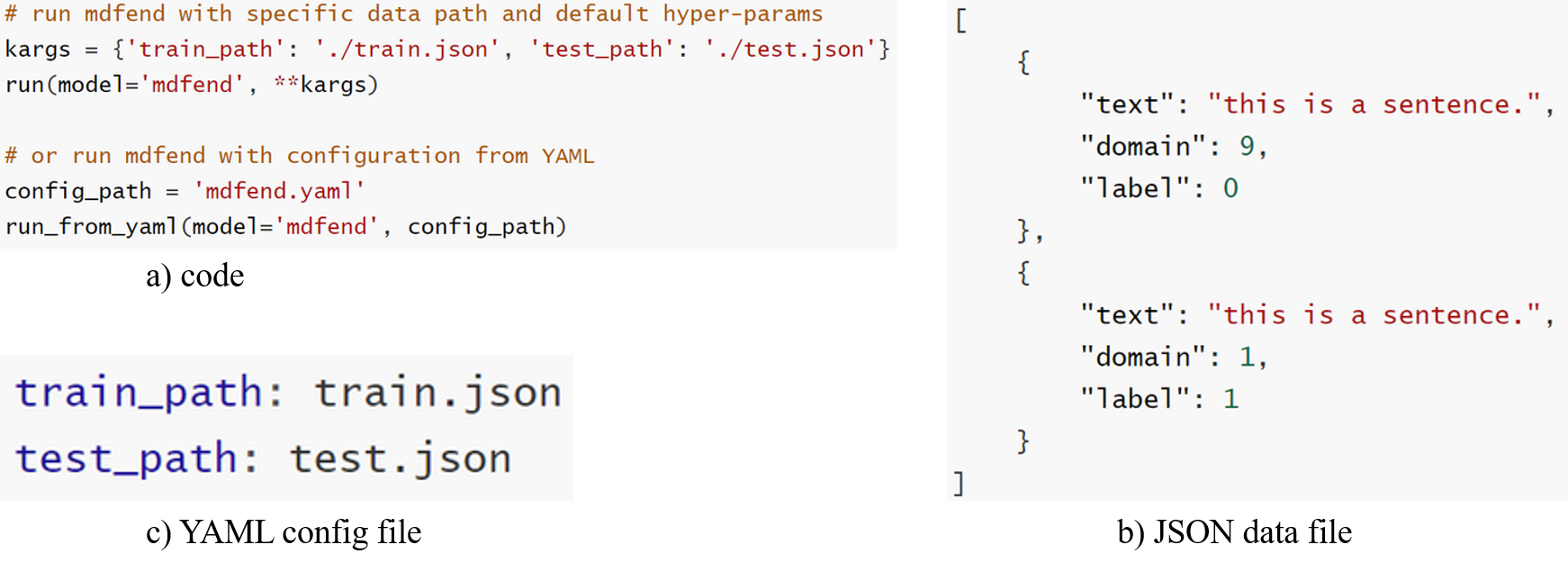}
    \caption{Quick start}
    \label{fig:start}
\end{figure*}

Furthermore, in scenarios where users prefer not to configure the intricacies of model training, a simplified approach is available. By specifying only a few essential arguments, such as the dataset, users can delegate to FaKnow the responsibility of determining the default values for model initialization and hyper-parameters, including the learning rate. These default values are derived from the relevant specifications provided in the open-source code of the respective paper.

In addition, the source code of the \textit{run} function itself is also a good example for users to run the various integrated models in FaKnow. The code in \textit{run} is not refactored with additional abstractions on purpose so that researchers can quickly iterate on each of the models without diving into additional abstractions or files.

Figure \ref{fig:start} shows an example of the MDFEND\cite{mdfend}(mentioned in Sec \ref{sec:fnd}) model using the \textit{run} and \textit{run\_from\_yaml} functions, respectively, both of which specify that the model name to be used is ``mdfend". The \textit{run} function specifies, via the keyword arguments, the paths to JSON files of the training set and test set, namely ``./train.json" and ``./test.json". JSON data files contain a list of key-value pairs of multiple samples, each with three attributes: text, domain, and label, which are required for training this model. The \textit{run\_from\_yaml} function requires the path to the YAML configuration file provided by the user, which also indicates the paths to JSON data files via key-value pairs. In this example, neither the validation set path nor the hyper-parameters are specified, so FaKnow will use the hyper-parameters in the code released with the paper to train and test MDFEND model.

\subsubsection{Train from scratch}

\begin{figure}[!htbp]
    \centering
    \includegraphics[width=0.49\textwidth]{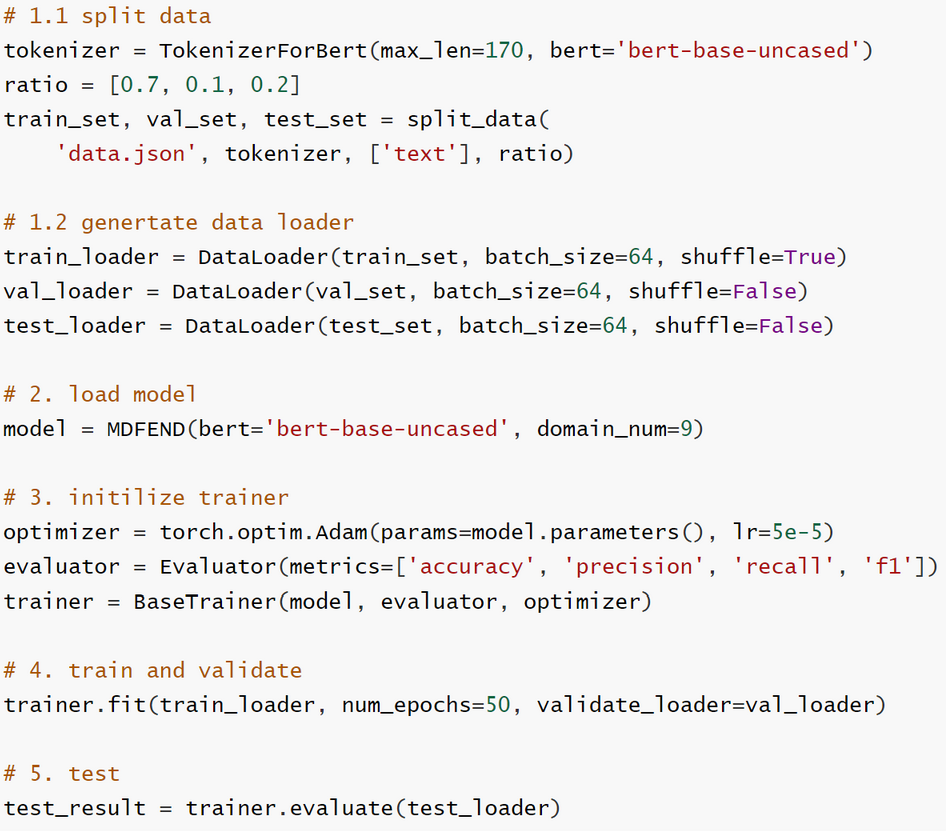}
    \caption{Train from scratch}
    \label{fig:scratch}
\end{figure}

As shown in the example in Figure \ref{fig:scratch}, to exercise complete control over the training process, users have the option of utilizing FaKnow to construct code for model training and evaluation right from the ground up. The specific steps involved are elaborated below.

\paragraph{Prepare Data}
Users should generate the \textit{pytorch.Dataloader} for data to be used. FaKnow offers a comprehensive set of \textit{Dataset} classes for the built-in models, accompanied by a diverse range of data processing functionalities (e.g., text segmentation, image conversion, etc.). Alternatively, users may opt to utilize customized \textit{Dataset} classes. In this example, the tokenizer with a maximum text length of 170 and uncased base BERT is generated to tokenize the text field in the JSON data file.
Then, the data in the ``data.json" file is proportionally divided into the training set, validation set, and test set by the \textit{split\_data} function, and corresponding data loaders with a batch size of 64 are created for each of them.

\paragraph{Load Model}
This step involves the loading of the intended model for the training process. In the example code, an MDFEND class with uncased base BERT is instantiated, and the number of news domains is specified as 9.

\paragraph{Initialize Trainer}
Users must initialize the \textit{Trainer} responsible for model training. This entails choosing an appropriate optimization algorithm and defining the evaluation metrics. Furthermore, additional advanced settings are supported here, such as learning rate schedulers, gradient clipping, early stopping, and the device for training. In this code, accuracy, precision, recall, and f1-score are taken as evaluation metrics to generate an evaluator and Adam is specified as the optimizer with a learning rate of 0.00005. Here, \textit{BaseTrainer} class is initialized, which is a subclass of \textit{Trainer} and can adapt to the majority of circumstances.

\paragraph{Train and validate}
The \textit{fit} method of \textit{Trainer} is called to execute the training and validation for the model and the results are subsequently saved. If the validation set was not created during the data preparation stage, only model training will be conducted. In this example, the trainer will train and validate for a duration of 50 epochs on the previously generated data loaders.

\paragraph{Test}
 By invoking the \textit{evaluate} method of \textit{Trainer}, users can assess the model's performance on the test set, based on the evaluation metrics set during the third step. In Figure \ref{fig:scratch}, the accuracy, precision, recall, and f1-score previously specified through the evaluator will be returned as test results.

\subsection{Develop New Models} \label{sec:develop}
\begin{figure}[!htbp]
    \centering
    \includegraphics[width=0.5\textwidth]{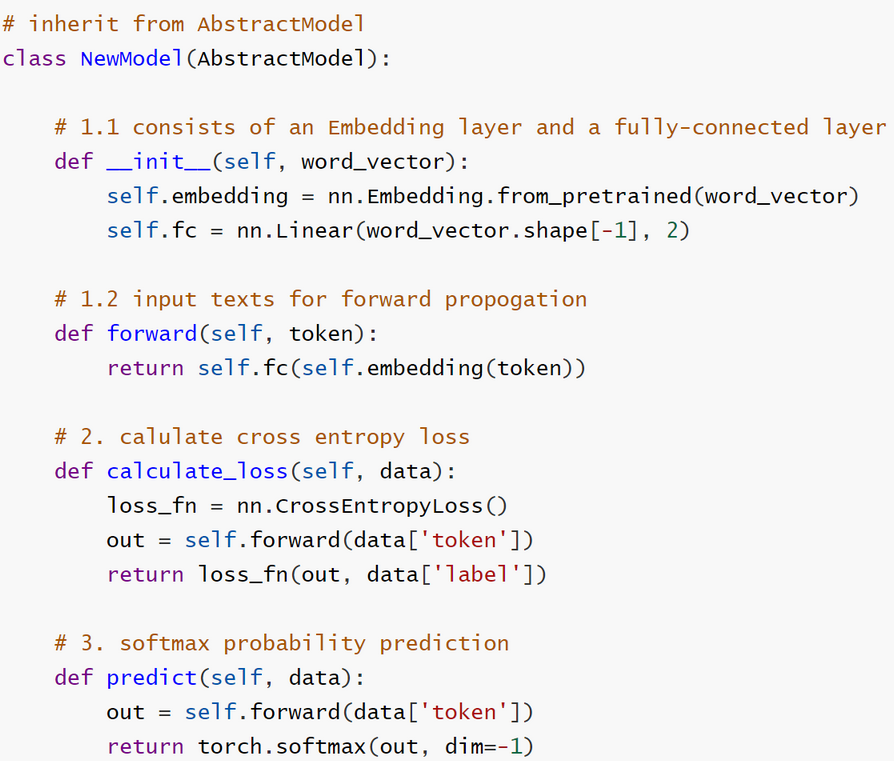}
    \caption{Develop new models}
    \label{fig:new}
\end{figure}

As detailed in Sec \ref{sec:interface}, we have formulated a comprehensive interface for all models integrated within FaKnow. The new model developed by users should inherit from \textit{AbstractModel} and override the corresponding methods as outlined in the ensuing steps. Figure \ref{fig:new} also illustrates an example of developing a simple model with a word embedding layer and a fully connected layer, which only uses the text in the post for detection.

\paragraph{Implement \textit{\_\_init\_\_} and \textit{forward}}
Since all models indirectly inherit from the \textit{nn.Module} within PyTorch(shown in Figure \ref{fig:abstractmodel}), the way of overriding the \textit{\_\_init\_\_} and \textit{forward} replicates the standard methodology employed while utilizing PyTorch directly. Within the \textit{\_\_init\_\_} method, various parameters are initialized and member variables relevant to the model are defined. Conversely, \textit{forward} necessitates the completion of forward propagation, encompassing the reception of an input batch comprising sample data, culminating in the generation of the output from the model's final layer. In this example, an embedding layer from pre-trained word vectors and a fully connected layer for text classification are defined in the \textit{\_\_init\_\_} method. Then the input text tokens are passed through these two layers in turn to get the final output of the model in the \textit{forward} method.

\paragraph{Implement \textit{calculate\_loss}}
As shown in Figure \ref{fig:interface}, users are expected to compose the logic code that facilitates the calculation of loss within this method. It entails invoking \textit{forward} to acquire the output from the model's final layer and performing the loss computation based on the ground truth associated with the samples. In scenarios where the final loss entails multiple losses, the user can also construct a \textit{python.Dict} to collectively return them. In Figure \ref{fig:new}, the text tokens and labels are obtained from the dict batch data mentioned in Sec \ref{sec:dict_batch_data} according to the corresponding key respectively, and the cross-entropy is employed as the loss function to return the final loss.

\paragraph{Implement \textit{predict}}
Derived from the output of the \textit{forward} method, users are required to return the probability of given batch samples being classified as either true or fake news. In this code, the tokens are also retrieved from the dictionary batch data, and the sotfmax prediction is returned based on the model's output.

\section{Conclusion}\label{sec:conclusion}
In this paper, we introduce FaKnow, a comprehensive library comprising a collection of fake news detection models that encompass two major categories: content-based and social context-based. FaKnow is designed to offer a unified framework for these algorithms, encompassing a sequence of processes such as data processing, model training, and evaluation, as well as supplementary functionalities such as visualization and logging. With these functionalities and PyTorch-based logic behaviors, it offers a user-friendly and seamless initiation process, ensuring a delightful interactive experience. Furthermore, a carefully designed and standardized API ensures excellent extensibility of the library, empowering users to effortlessly customize diverse functions for specific scenarios with minimal code requirements.

By providing this cohesive framework, FaKnow contributes to the harmonization of research efforts in fake news detection, enabling subsequent researchers to effortlessly replicate existing algorithms or develop new models. In the future, we will commit to adding new models to FaKnow and developing new features to facilitate the usage of the library continually.

\section*{Author Contribution}
Yongjun Li proposed the idea of this work and gave many helpful suggestions. Yiyuan Zhu designed the research framework, reproduced most of the integrated models, and wrote the paper. Jialiang Wang, Ming Gao, and Jiali Wei reproduced the remaining models. In addition, the documentation of our open-source library was written by Jialiang Wang.

\section*{Acknowledgment}
This work is supported by 
Key Research and Development Program in Shaanxi Province of China (Program No. 2024GX-YBXM-124).

\printbibliography

@inproceedings{bert,
  title = {BERT: Pre-Training of Deep Bidirectional Transformers for Language Understanding},
  shorttitle = {BERT},
  booktitle = {Proceedings of the 2019 Conference of the North American Chapter of the Association for Computational Linguistics: Human Language Technologies, Volume 1 (Long and Short Papers)},
  author = {Devlin, Jacob and Chang, Ming-Wei and Lee, Kenton and Toutanova, Kristina},
  date = {2019-06},
  pages = {4171--4186},
  publisher = {Association for Computational Linguistics},
  location = {Minneapolis, Minnesota},
  doi = {10.18653/v1/N19-1423},
  url = {https://aclanthology.org/N19-1423},
  eventtitle = {NAACL-HLT 2019}
}

@inproceedings{bigcn,
  title = {Rumor Detection on Social Media with Bi-Directional Graph Convolutional Networks},
  booktitle = {Proceedings of the AAAI Conference on Artificial Intelligence},
  author = {Bian, Tian and Xiao, Xi and Xu, Tingyang and Zhao, Peilin and Huang, Wenbing and Rong, Yu and Huang, Junzhou},
  date = {2020-04-03},
  volume = {34},
  pages = {549--556},
  doi = {10.1609/aaai.v34i01.5393},
  url = {https://ojs.aaai.org/index.php/AAAI/article/view/5393},
  langid = {english}
}

@inproceedings{cafe,
  title = {Cross-Modal Ambiguity Learning for Multimodal Fake News Detection},
  booktitle = {Proceedings of the ACM Web Conference 2022},
  author = {Chen, Yixuan and Li, Dongsheng and Zhang, Peng and Sui, Jie and Lv, Qin and Tun, Lu and Shang, Li},
  date = {2022-04-25},
  pages = {2897--2905},
  publisher = {ACM},
  location = {Virtual Event, Lyon France},
  doi = {10.1145/3485447.3511968},
  url = {https://dl.acm.org/doi/10.1145/3485447.3511968},
  eventtitle = {WWW '22: The ACM Web Conference 2022},
  isbn = {978-1-4503-9096-5},
  langid = {english}
}

@article{ced,
  title = {CED: Credible Early Detection of Social Media Rumors},
  shorttitle = {CED},
  author = {Song, Changhe and Yang, Cheng and Chen, Huimin and Tu, Cunchao and Liu, Zhiyuan and Sun, Maosong},
  date = {2021-08},
  journaltitle = {IEEE Transactions on Knowledge and Data Engineering},
  volume = {33},
  number = {8},
  pages = {3035--3047},
  issn = {1558-2191},
  doi = {10.1109/TKDE.2019.2961675},
  url = {https://ieeexplore.ieee.org/document/8939421},
  eventtitle = {IEEE Transactions on Knowledge and Data Engineering}
}

@inproceedings{chengVRoCVariationalAutoencoderaided2020,
  title = {VRoC: Variational Autoencoder-Aided Multi-Task Rumor Classifier Based on Text},
  shorttitle = {VRoC},
  booktitle = {Proceedings of The Web Conference 2020},
  author = {Cheng, Mingxi and Nazarian, Shahin and Bogdan, Paul},
  date = {2020-04-20},
  pages = {2892--2898},
  publisher = {ACM},
  location = {Taipei Taiwan},
  doi = {10.1145/3366423.3380054},
  url = {https://dl.acm.org/doi/10.1145/3366423.3380054},
  eventtitle = {WWW '20: The Web Conference 2020},
  isbn = {9781450370233},
  langid = {english}
}

@article{csfnd,
  title = {Not All Fake News Is Semantically Similar: Contextual Semantic Representation Learning for Multimodal Fake News Detection},
  shorttitle = {Not All Fake News Is Semantically Similar},
  author = {Peng, Liwen and Jian, Songlei and Kan, Zhigang and Qiao, Linbo and Li, Dongsheng},
  date = {2024-01-01},
  journaltitle = {Information Processing \& Management},
  shortjournal = {Information Processing \& Management},
  volume = {61},
  number = {1},
  pages = {103564},
  issn = {0306-4573},
  doi = {10.1016/j.ipm.2023.103564},
  url = {https://www.sciencedirect.com/science/article/pii/S0306457323003011}
}

@inproceedings{dudef,
  title = {Mining Dual Emotion for Fake News Detection},
  booktitle = {Proceedings of the Web Conference 2021},
  author = {Zhang, Xueyao and Cao, Juan and Li, Xirong and Sheng, Qiang and Zhong, Lei and Shu, Kai},
  date = {2021-06-03},
  series = {WWW '21},
  pages = {3465--3476},
  publisher = {Association for Computing Machinery},
  location = {New York, NY, USA},
  doi = {10.1145/3442381.3450004},
  url = {https://dl.acm.org/doi/10.1145/3442381.3450004},
  isbn = {978-1-4503-8312-7}
}

@inproceedings{eann,
  title = {EANN: Event Adversarial Neural Networks for Multi-Modal Fake News Detection},
  booktitle = {Proceedings of the 24th ACM SIGKDD International Conference on Knowledge Discovery \& Data Mining},
  author = {Wang, Yaqing and Ma, Fenglong and Jin, Zhiwei and Yuan, Ye and Xun, Guangxu and Jha, Kishlay and Su, Lu and Gao, Jing},
  date = {2018},
  url = {https://api.semanticscholar.org/CorpusID:46990556}
}

@inproceedings{ebgcn,
  title = {Towards Propagation Uncertainty: Edge-Enhanced Bayesian Graph Convolutional Networks for Rumor Detection},
  shorttitle = {Towards Propagation Uncertainty},
  booktitle = {Proceedings of the 59th Annual Meeting of the Association for Computational Linguistics and the 11th International Joint Conference on Natural Language Processing (Volume 1: Long Papers)},
  author = {Wei, Lingwei and Hu, Dou and Zhou, Wei and Yue, Zhaojuan and Hu, Songlin},
  date = {2021},
  pages = {3845--3854},
  publisher = {Association for Computational Linguistics},
  location = {Online},
  doi = {10.18653/v1/2021.acl-long.297},
  url = {https://aclanthology.org/2021.acl-long.297},
  eventtitle = {Proceedings of the 59th Annual Meeting of the Association for Computational Linguistics and the 11th International Joint Conference on Natural Language Processing (Volume 1: Long Papers)},
  langid = {english}
}

@inproceedings{endfn,
  title = {Generalizing to the Future: Mitigating Entity Bias in Fake News Detection},
  shorttitle = {Generalizing to the Future},
  booktitle = {Proceedings of the 45th International ACM SIGIR Conference on Research and Development in Information Retrieval},
  author = {Zhu, Yongchun and Sheng, Qiang and Cao, Juan and Li, Shuokai and Wang, Danding and Zhuang, Fuzhen},
  date = {2022-07-07},
  series = {SIGIR '22},
  pages = {2120--2125},
  publisher = {Association for Computing Machinery},
  location = {New York, NY, USA},
  doi = {10.1145/3477495.3531816},
  url = {https://dl.acm.org/doi/10.1145/3477495.3531816},
  isbn = {978-1-4503-8732-3}
}

@unpublished{fakenewsnet,
  title = {FakeNewsNet: A Data Repository with News Content, Social Context and Dynamic Information for Studying Fake News on Social Media},
  author = {Shu, Kai and Mahudeswaran, Deepak and Wang, Suhang and Lee, Dongwon and Liu, Huan},
  date = {2018},
  eprint = {1809.01286},
  eprinttype = {arxiv}
}

@inproceedings{fang,
  title = {FANG: Leveraging Social Context for Fake News Detection Using Graph Representation},
  shorttitle = {FANG},
  booktitle = {Proceedings of the 29th ACM International Conference on Information \& Knowledge Management},
  author = {Nguyen, Van-Hoang and Sugiyama, Kazunari and Nakov, Preslav and Kan, Min-Yen},
  date = {2020-10-19},
  series = {CIKM '20},
  pages = {1165--1174},
  publisher = {Association for Computing Machinery},
  location = {New York, NY, USA},
  doi = {10.1145/3340531.3412046},
  url = {https://dl.acm.org/doi/10.1145/3340531.3412046},
  isbn = {978-1-4503-6859-9}
}

@inproceedings{gat,
  title = {Graph Attention Networks},
  author = {Veličković, Petar and Cucurull, Guillem and Casanova, Arantxa and Romero, Adriana and Liò, Pietro and Bengio, Yoshua},
  date = {2018-02-15},
  url = {https://openreview.net/forum?id=rJXMpikCZ},
  eventtitle = {International Conference on Learning Representations},
  langid = {english}
}

@inproceedings{gcan,
  title = {GCAN: Graph-Aware Co-Attention Networks for Explainable Fake News Detection on Social Media},
  shorttitle = {GCAN},
  booktitle = {Proceedings of the 58th Annual Meeting of the Association for Computational Linguistics},
  author = {Lu, Yi-Ju and Li, Cheng-Te},
  date = {2020-07},
  pages = {505--514},
  publisher = {Association for Computational Linguistics},
  location = {Online},
  doi = {10.18653/v1/2020.acl-main.48},
  url = {https://aclanthology.org/2020.acl-main.48},
  eventtitle = {ACL 2020}
}

@inproceedings{gcn,
  title = {Semi-Supervised Classification with Graph Convolutional Networks},
  author = {Kipf, Thomas N. and Welling, Max},
  date = {2016-11-03},
  url = {https://openreview.net/forum?id=SJU4ayYgl},
  eventtitle = {International Conference on Learning Representations},
  langid = {english}
}

@article{gcnfn,
  title = {Fake News Detection on Social Media Using Geometric Deep Learning},
  author = {Monti, Federico and Frasca, Fabrizio and Eynard, Davide and Mannion, Damon and Bronstein, Michael M.},
  date = {2019},
  journaltitle = {ArXiv},
  volume = {abs/1902.06673},
  url = {https://api.semanticscholar.org/CorpusID:62841478}
}

@inproceedings{gnncl,
  title = {Continual Learning for Fake News Detection from Social Media},
  booktitle = {Artificial Neural Networks and Machine Learning – ICANN 2021: 30th International Conference on Artificial Neural Networks, Bratislava, Slovakia, September 14–17, 2021, Proceedings, Part II},
  author = {Han, Yi and Karunasekera, Shanika and Leckie, Christopher},
  date = {2021-09-14},
  pages = {372--384},
  publisher = {Springer-Verlag},
  location = {Berlin, Heidelberg},
  doi = {10.1007/978-3-030-86340-1_30},
  url = {https://doi.org/10.1007/978-3-030-86340-1_30},
  isbn = {978-3-030-86339-5}
}

@inproceedings{graphsage,
  title = {Inductive Representation Learning on Large Graphs},
  booktitle = {Proceedings of the 31st International Conference on Neural Information Processing Systems},
  author = {Hamilton, William L. and Ying, Rex and Leskovec, Jure},
  date = {2017},
  series = {NIPS'17},
  pages = {1025--1035},
  publisher = {Curran Associates Inc.},
  location = {Red Hook, NY, USA},
  isbn = {978-1-5108-6096-4},
  venue = {Long Beach, California, USA}
}

@inproceedings{hmcan,
  title = {Hierarchical Multi-Modal Contextual Attention Network for Fake News Detection},
  booktitle = {Proceedings of the 44th International ACM SIGIR Conference on Research and Development in Information Retrieval},
  author = {Qian, Shengsheng and Wang, Jinguang and Hu, Jun and Fang, Quan and Xu, Changsheng},
  date = {2021},
  series = {SIGIR '21},
  pages = {153--162},
  publisher = {Association for Computing Machinery},
  location = {New York, NY, USA},
  doi = {10.1145/3404835.3462871},
  url = {https://doi.org/10.1145/3404835.3462871},
  isbn = {978-1-4503-8037-9},
  venue = {Virtual Event, Canada}
}

@inproceedings{libcity,
  title = {LibCity: An Open Library for Traffic Prediction},
  booktitle = {Proceedings of the 29th International Conference on Advances in Geographic Information Systems},
  author = {Wang, Jingyuan and Jiang, Jiawei and Jiang, Wenjun and Li, Chao and Zhao, Wayne Xin},
  date = {2021},
  series = {SIGSPATIAL '21},
  pages = {145--148},
  publisher = {Association for Computing Machinery},
  location = {New York, NY, USA},
  doi = {10.1145/3474717.3483923},
  url = {https://doi.org/10.1145/3474717.3483923},
  isbn = {978-1-4503-8664-7},
  pagetotal = {4}
}

@article{lstm,
  title = {Long Short-Term Memory},
  author = {Hochreiter, Sepp and Schmidhuber, Jürgen},
  date = {1997-11},
  journaltitle = {Neural Comput.},
  volume = {9},
  number = {8},
  pages = {1735--1780},
  publisher = {MIT Press},
  location = {Cambridge, MA, USA},
  issn = {0899-7667},
  doi = {10.1162/neco.1997.9.8.1735},
  url = {https://doi.org/10.1162/neco.1997.9.8.1735}
}

@article{m3fend,
  title = {Memory-Guided Multi-View Multi-Domain Fake News Detection},
  author = {Zhu, Yongchun and Sheng, Qiang and Cao, Juan and Nan, Qiong and Shu, Kai and Wu, Minghui and Wang, Jindong and Zhuang, Fuzhen},
  date = {2023-07},
  journaltitle = {IEEE Transactions on Knowledge and Data Engineering},
  volume = {35},
  number = {7},
  pages = {7178--7191},
  issn = {1558-2191},
  doi = {10.1109/TKDE.2022.3185151},
  url = {https://ieeexplore.ieee.org/document/9802916},
  eventtitle = {IEEE Transactions on Knowledge and Data Engineering}
}

@inproceedings{mcan,
  title = {Multimodal Fusion with Co-Attention Networks for Fake News Detection},
  booktitle = {Findings of the Association for Computational Linguistics: ACL-IJCNLP 2021},
  author = {Wu, Yang and Zhan, Pengwei and Zhang, Yunjian and Wang, Liming and Xu, Zhen},
  date = {2021-08},
  pages = {2560--2569},
  publisher = {Association for Computational Linguistics},
  location = {Online},
  doi = {10.18653/v1/2021.findings-acl.226},
  url = {https://aclanthology.org/2021.findings-acl.226},
  eventtitle = {Findings 2021}
}

@inproceedings{mdfend,
  title = {MDFEND: Multi-Domain Fake News Detection},
  booktitle = {Proceedings of the 30th ACM International Conference on Information \& Knowledge Management},
  author = {Nan, Qiong and Cao, Juan and Zhu, Yongchun and Wang, Yanyan and Li, Jintao},
  date = {2021},
  series = {CIKM '21},
  pages = {3343--3347},
  publisher = {Association for Computing Machinery},
  location = {New York, NY, USA},
  doi = {10.1145/3459637.3482139},
  url = {https://doi.org/10.1145/3459637.3482139},
  isbn = {978-1-4503-8446-9},
  pagetotal = {5}
}

@article{mediaeval1516,
  title = {Detection and Visualization of Misleading Content on Twitter},
  author = {Boididou, Christina and Papadopoulos, Symeon and Zampoglou, Markos and Apostolidis, Lazaros and Papadopoulou, Olga and Kompatsiaris, Yiannis},
  date = {2018-03-01},
  journaltitle = {International Journal of Multimedia Information Retrieval},
  shortjournal = {Int J Multimed Info Retr},
  volume = {7},
  number = {1},
  pages = {71--86},
  issn = {2192-662X},
  doi = {10.1007/s13735-017-0143-x},
  url = {https://doi.org/10.1007/s13735-017-0143-x},
  langid = {english}
}

@online{mmrec,
  title = {MMRec: Simplifying Multimodal Recommendation},
  shorttitle = {MMRec},
  author = {Zhou, Xin},
  date = {2023-02-02},
  url = {https://arxiv.org/abs/2302.03497v1},
  langid = {english},
  organization = {arXiv.org}
}

@article{pheme,
  title = {PHEME Dataset for Rumour Detection and Veracity Classification},
  author = {Kochkina, Elena and Liakata, Maria and Zubiaga, Arkaitz},
  date = {2018-06},
  doi = {10.6084/m9.figshare.6392078.v1},
  url = {https://figshare.com/articles/dataset/PHEME_dataset_for_Rumour_Detection_and_Veracity_Classification/6392078}
}

@inproceedings{emfend,
  title = {Improving Fake News Detection by Using an Entity-Enhanced Framework to Fuse Diverse Multimodal Clues},
  booktitle = {Proceedings of the 29th ACM International Conference on Multimedia},
  author = {Qi, Peng and Cao, Juan and Li, Xirong and Liu, Huan and Sheng, Qiang and Mi, Xiaoyue and He, Qin and Lv, Yongbiao and Guo, Chenyang and Yu, Yingchao},
  date = {2021},
  series = {MM '21},
  pages = {1212--1220},
  publisher = {Association for Computing Machinery},
  location = {New York, NY, USA},
  doi = {10.1145/3474085.3481548},
  url = {https://doi.org/10.1145/3474085.3481548},
  isbn = {978-1-4503-8651-7},
  venue = {Virtual Event, China}
}

@inproceedings{recbole,
  title = {RecBole: Towards a Unified, Comprehensive and Efficient Framework for Recommendation Algorithms},
  booktitle = {Proceedings of the 30th ACM International Conference on Information \& Knowledge Management},
  author = {Zhao, Wayne Xin and Mu, Shanlei and Hou, Yupeng and Lin, Zihan and Chen, Yushuo and Pan, Xingyu and Li, Kaiyuan and Lu, Yujie and Wang, Hui and Tian, Changxin and Min, Yingqian and Feng, Zhichao and Fan, Xinyan and Chen, Xu and Wang, Pengfei and Ji, Wendi and Li, Yaliang and Wang, Xiaoling and Wen, Ji-Rong},
  date = {2021},
  series = {CIKM '21},
  pages = {4653--4664},
  publisher = {Association for Computing Machinery},
  location = {New York, NY, USA},
  doi = {10.1145/3459637.3482016},
  url = {https://doi.org/10.1145/3459637.3482016},
  isbn = {978-1-4503-8446-9},
  pagetotal = {12}
}

@inproceedings{safe,
  title = {SAFE: Similarity-Aware Multi-Modal Fake News Detection},
  shorttitle = {\$\$\textbackslash mathsf \{SAFE\}\$\$},
  booktitle = {Advances in Knowledge Discovery and Data Mining},
  author = {Zhou, Xinyi and Wu, Jindi and Zafarani, Reza},
  editor = {Lauw, Hady W. and Wong, Raymond Chi-Wing and Ntoulas, Alexandros and Lim, Ee-Peng and Ng, See-Kiong and Pan, Sinno Jialin},
  date = {2020},
  series = {Lecture Notes in Computer Science},
  pages = {354--367},
  publisher = {Springer International Publishing},
  location = {Cham},
  doi = {10.1007/978-3-030-47436-2_27},
  isbn = {978-3-030-47436-2},
  langid = {english}
}

@inproceedings{spotfake,
  title = {SpotFake: A Multi-Modal Framework for Fake News Detection},
  shorttitle = {SpotFake},
  booktitle = {2019 IEEE Fifth International Conference on Multimedia Big Data (BigMM)},
  author = {Singhal, Shivangi and Shah, Rajiv Ratn and Chakraborty, Tanmoy and Kumaraguru, Ponnurangam and Satoh, Shin'ichi},
  date = {2019-09},
  pages = {39--47},
  doi = {10.1109/BigMM.2019.00-44},
  eventtitle = {2019 IEEE Fifth International Conference on Multimedia Big Data (BigMM)}
}

@inproceedings{textcnn,
  title = {Convolutional Neural Networks for Sentence Classification},
  booktitle = {Proceedings of the 2014 Conference on Empirical Methods in Natural Language Processing (EMNLP)},
  author = {Kim, Yoon},
  date = {2014-10},
  pages = {1746--1751},
  publisher = {Association for Computational Linguistics},
  location = {Doha, Qatar},
  doi = {10.3115/v1/D14-1181},
  url = {https://aclanthology.org/D14-1181},
  eventtitle = {EMNLP 2014}
}

@inproceedings{transformers,
  title = {Transformers: State-of-the-Art Natural Language Processing},
  shorttitle = {Transformers},
  booktitle = {Proceedings of the 2020 Conference on Empirical Methods in Natural Language Processing: System Demonstrations},
  author = {Wolf, Thomas and Debut, Lysandre and Sanh, Victor and Chaumond, Julien and Delangue, Clement and Moi, Anthony and Cistac, Pierric and Rault, Tim and Louf, Remi and Funtowicz, Morgan and Davison, Joe and Shleifer, Sam and family=Platen, given=Patrick, prefix=von, useprefix=true and Ma, Clara and Jernite, Yacine and Plu, Julien and Xu, Canwen and Le Scao, Teven and Gugger, Sylvain and Drame, Mariama and Lhoest, Quentin and Rush, Alexander},
  editor = {Liu, Qun and Schlangen, David},
  date = {2020-10},
  pages = {38--45},
  publisher = {Association for Computational Linguistics},
  location = {Online},
  doi = {10.18653/v1/2020.emnlp-demos.6},
  url = {https://aclanthology.org/2020.emnlp-demos.6}
}

@inproceedings{trustrd,
  title = {Towards Trustworthy Rumor Detection with Interpretable Graph Structural Learning},
  booktitle = {Proceedings of the 32nd ACM International Conference on Information and Knowledge Management},
  author = {Liu, Leyuan and Chen, Junyi and Cheng, Zhangtao and Tai, Wenxin and Zhou, Fan},
  date = {2023-10-21},
  series = {CIKM '23},
  pages = {4089--4093},
  publisher = {Association for Computing Machinery},
  location = {New York, NY, USA},
  doi = {10.1145/3583780.3615228},
  url = {https://dl.acm.org/doi/10.1145/3583780.3615228}
}

@inproceedings{twitter1516,
  title = {Detect Rumors in Microblog Posts Using Propagation Structure via Kernel Learning},
  booktitle = {Proceedings of the 55th Annual Meeting of the Association for Computational Linguistics (Volume 1: Long Papers)},
  author = {Ma, Jing and Gao, Wei and Wong, Kam-Fai},
  editor = {Barzilay, Regina and Kan, Min-Yen},
  date = {2017-07},
  pages = {708--717},
  publisher = {Association for Computational Linguistics},
  location = {Vancouver, Canada},
  doi = {10.18653/v1/P17-1066},
  url = {https://aclanthology.org/P17-1066},
  eventtitle = {ACL 2017}
}

@inproceedings{upfd,
  title = {User Preference-Aware Fake News Detection},
  booktitle = {Proceedings of the 44th International ACM SIGIR Conference on Research and Development in Information Retrieval},
  author = {Dou, Yingtong and Shu, Kai and Xia, Congying and Yu, Philip S. and Sun, Lichao},
  date = {2021},
  series = {SIGIR '21},
  pages = {2051--2055},
  publisher = {Association for Computing Machinery},
  location = {New York, NY, USA},
  doi = {10.1145/3404835.3462990},
  url = {https://doi.org/10.1145/3404835.3462990},
  isbn = {978-1-4503-8037-9},
  pagetotal = {5}
}

@inproceedings{vaibhavSentenceInteractionsMatter2019,
  title = {Do Sentence Interactions Matter? Leveraging Sentence Level Representations for Fake News Classification},
  shorttitle = {Do Sentence Interactions Matter?},
  booktitle = {Proceedings of the Thirteenth Workshop on Graph-Based Methods for Natural Language Processing (TextGraphs-13)},
  author = {Vaibhav, Vaibhav and Mandyam, Raghuram and Hovy, Eduard},
  date = {2019-11},
  pages = {134--139},
  publisher = {Association for Computational Linguistics},
  location = {Hong Kong},
  doi = {10.18653/v1/D19-5316},
  url = {https://aclanthology.org/D19-5316},
  eventtitle = {TextGraphs 2019}
}

@article{w2v,
  title = {Efficient Estimation of Word Representations in Vector Space},
  author = {Mikolov, Tomas and Chen, Kai and Corrado, G.s and Dean, Jeffrey},
  date = {2013-01},
  journaltitle = {Proceedings of Workshop at ICLR},
  volume = {2013}
}

@inproceedings{weibo17,
  title = {Multimodal Fusion with Recurrent Neural Networks for Rumor Detection on Microblogs},
  booktitle = {Proceedings of the 25th ACM International Conference on Multimedia},
  author = {Jin, Zhiwei and Cao, Juan and Guo, Han and Zhang, Yongdong and Luo, Jiebo},
  date = {2017-10-19},
  series = {MM '17},
  pages = {795--816},
  publisher = {Association for Computing Machinery},
  location = {New York, NY, USA},
  doi = {10.1145/3123266.3123454},
  url = {https://dl.acm.org/doi/10.1145/3123266.3123454},
  isbn = {978-1-4503-4906-2}
}

@inproceedings{yuConvolutionalApproachMisinformation2017a,
  title = {A Convolutional Approach for Misinformation Identification},
  booktitle = {Proceedings of the Twenty-Sixth International Joint Conference on Artificial Intelligence},
  author = {Yu, Feng and Liu, Qiang and Wu, Shu and Wang, Liang and Tan, Tieniu},
  date = {2017-08},
  pages = {3901--3907},
  publisher = {International Joint Conferences on Artificial Intelligence Organization},
  location = {Melbourne, Australia},
  doi = {10.24963/ijcai.2017/545},
  url = {https://www.ijcai.org/proceedings/2017/545},
  eventtitle = {Twenty-Sixth International Joint Conference on Artificial Intelligence},
  isbn = {978-0-9992411-0-3},
  langid = {english}
}

@inproceedings{vgg,
  title = {Very Deep Convolutional Networks for Large-Scale Image Recognition},
  booktitle = {3rd International Conference on Learning Representations, ICLR 2015, San Diego, CA, USA, May 7-9, 2015, Conference Track Proceedings},
  author = {Simonyan, Karen and Zisserman, Andrew},
  editor = {Bengio, Yoshua and LeCun, Yann},
  date = {2015},
  url = {http://arxiv.org/abs/1409.1556},
  bibsource = {dblp computer science bibliography, https://dblp.org}
}

@inproceedings{mfan,
  title = {MFAN: Multi-Modal Feature-Enhanced Attention Networks for Rumor Detection},
  shorttitle = {MFAN},
  author = {Zheng, Jiaqi and Zhang, Xi and Guo, Sanchuan and Wang, Quan and Zang, Wenyu and Zhang, Yongdong},
  date = {2022-07-16},
  volume = {3},
  pages = {2413--2419},
  issn = {1045-0823},
  doi = {10.24963/ijcai.2022/335},
  url = {https://www.ijcai.org/proceedings/2022/335},
  eventtitle = {Thirty-First International Joint Conference on Artificial Intelligence},
  langid = {english}
}

@inproceedings{nep,
  title = {Zoom Out and Observe: News Environment Perception for Fake News Detection},
  shorttitle = {Zoom Out and Observe},
  booktitle = {Proceedings of the 60th Annual Meeting of the Association for Computational Linguistics (Volume 1: Long Papers)},
  author = {Sheng, Qiang and Cao, Juan and Zhang, Xueyao and Li, Rundong and Wang, Danding and Zhu, Yongchun},
  editor = {Muresan, Smaranda and Nakov, Preslav and Villavicencio, Aline},
  date = {2022-05},
  pages = {4543--4556},
  publisher = {Association for Computational Linguistics},
  location = {Dublin, Ireland},
  doi = {10.18653/v1/2022.acl-long.311},
  url = {https://aclanthology.org/2022.acl-long.311},
  eventtitle = {ACL 2022}
}

@inproceedings{twitterma,
  title = {Detecting Rumors from Microblogs with Recurrent Neural Networks},
  booktitle = {Proceedings of the Twenty-Fifth International Joint Conference on Artificial Intelligence},
  author = {Ma, Jing and Gao, Wei and Mitra, Prasenjit and Kwon, Sejeong and Jansen, Bernard J. and Wong, Kam-Fai and Cha, Meeyoung},
  date = {2016},
  series = {IJCAI'16},
  pages = {3818--3824},
  publisher = {AAAI Press},
  location = {New York, New York, USA},
  isbn = {978-1-57735-770-4},
  pagetotal = {7}
}

@article{showandtell,
  title = {Show and Tell: Lessons Learned from the 2015 MSCOCO Image Captioning Challenge},
  shorttitle = {Show and Tell},
  author = {Vinyals, Oriol and Toshev, Alexander and Bengio, Samy and Erhan, Dumitru},
  date = {2017-04},
  journaltitle = {IEEE Transactions on Pattern Analysis and Machine Intelligence},
  volume = {39},
  number = {4},
  pages = {652--663},
  issn = {1939-3539},
  doi = {10.1109/TPAMI.2016.2587640},
  url = {https://ieeexplore.ieee.org/document/7505636},
  eventtitle = {IEEE Transactions on Pattern Analysis and Machine Intelligence}
}

@inproceedings{mmcovid,
  title = {Toward A Multilingual and Multimodal Data Repository for COVID-19 Disinformation},
  booktitle = {2020 IEEE International Conference on Big Data (Big Data)},
  author = {Li, Yichuan and Jiang, Bohan and Shu, Kai and Liu, Huan},
  date = {2020-12-10},
  pages = {4325--4330},
  publisher = {IEEE},
  location = {Atlanta, GA, USA},
  doi = {10.1109/BigData50022.2020.9378472},
  url = {https://ieeexplore.ieee.org/document/9378472/},
  eventtitle = {2020 IEEE International Conference on Big Data (Big Data)},
  isbn = {978-1-72816-251-5},
  langid = {english}
}

\end{document}